\documentclass[11pt]{article}
\usepackage{research-preprint}
\hypersetup{pdftitle={Retaining Repair Hypotheses for Cost-Aware Execution}}
\title{Retaining Repair Hypotheses for Cost-Aware Execution}
\author{Jae Hun Hwang\textsuperscript{1,2}\quad Hong-In Won\textsuperscript{1,3,*}\\[0.6em]
{\small\textsuperscript{1}Manufacturing AI Research Center, Korea Institute of Industrial Technology, Incheon, Republic of Korea}\\[0.25em]
{\small\textsuperscript{2}Department of Artificial Intelligence Convergence, Hanyang University, Republic of Korea}\\[0.25em]
{\small\textsuperscript{3}HYU-KITECH Joint Department, Graduate School, Hanyang University, Seoul, Republic of Korea}\\[0.4em]
{\small\textsuperscript{*}Corresponding author: \href{mailto:luvhayym@kitech.re.kr}{luvhayym@kitech.re.kr}}}

\date{Preprint}
\begin{document}\maketitle

\begin{abstract}

Several corrections of an executable model can fit the same observations while predicting different execution outcomes. We study a repair-to-execution interface that retains compatible finite guard repairs, conditions their weights on diagnostic and ordinary-execution observations, and uses their predictions to choose routes, priced tests, or a backup. On 24 synthetic two-stage devices, an equal mixture of a language-model-generated repair and a uniform belief over compatible repairs (the AI-weighted belief) lowers no-test cost by 2.14 percent relative to committing to the generated repair. Access charges can erase the benefit of diagnosis on isolated tasks, while repeated work can recover it under the AI-weighted belief. With normal-execution learning in every repeated-work arm, workload-valued testing costs 3.33 percent less than same-budget random selection under the AI-weighted belief. Weighting changes the comparison: uniform normal learning costs 540.75 less than AI-weighted normal learning on this cohort, with a device-paired interval excluding zero. Testing recovers part of that gap, but uniform normal learning still has a lower aggregate cost than AI-weighted workload-valued selection. Under uniform beliefs, information-gain selection has the lowest total among the compared selectors, while workload-valued testing costs more than normal learning. These results support retaining alternatives to a generated repair and evaluating diagnostic selection together with repair weighting and learning from execution.

\begin{figure}[H]\centering\includegraphics[width=\linewidth]{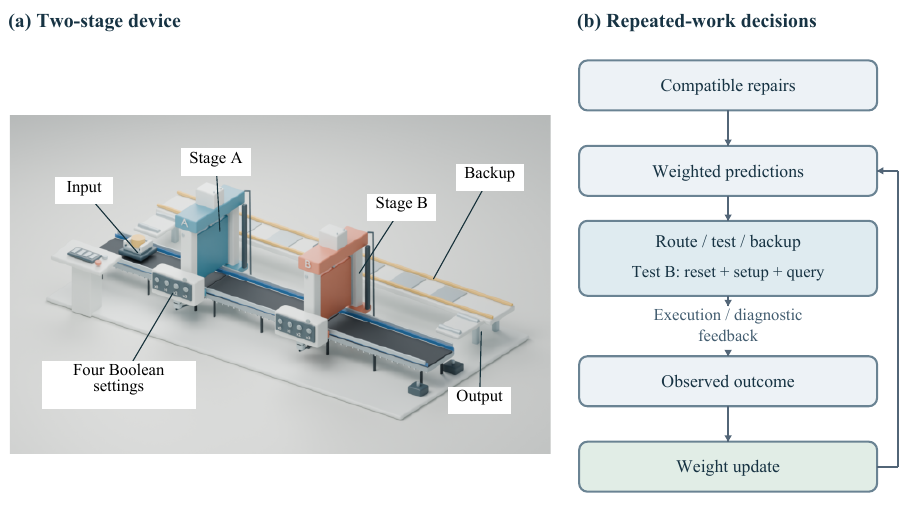}\caption{Compatible repairs turn model ambiguity into route, test and backup choices. (a) Illustrative synthetic device. (b) Repeated-work decision loop: tests and ordinary execution update the retained belief. Access-charged stage-B tests include reset (0.25), stage-A setup and query.}\end{figure}

\end{abstract}
\FloatBarrier
\section{Introduction}

Consider a controller using a program that predicts whether an operation is enabled. A failed operation reveals that the program is wrong, but usually does not identify a unique correction: one observation can be explained by several missing preconditions. Selecting one corrected program and treating it as certain can turn a successful local repair into another downstream failure.

Planning-domain learning and repair construct an action model that explains available evidence. FAMA compiles partially observed traces into a learning problem \citep{ref11}; hitting-set repair makes known-positive plans valid with a cardinality-minimal edit set \citep{ref12}, and counter-example repair also requires negative plans to remain invalid \citep{ref13}. These objectives belong to the broader model-repair landscape \citep{ref17}. Recent LLM-only repair work separately evaluates recovered model structure and satisfaction of test-plan constraints \citep{ref18}. This distinction matters because a locally consistent repair can still leave uncertainty about unobserved behavior.

Executable world-model methods connect model construction to acting: WorldCoder uses replay consistency and optimistic goal reachability \citep{ref4}, while Generate, Improve and Fix with Monte Carlo Tree Search (GIF-MCTS) searches generated code using observed transitions \citep{ref5}. LLM-SMC-S, a sequential Monte Carlo sampler over hypotheses proposed by a language model, maintains weighted hypotheses and directs new experiments \citep{ref10}. These methods establish useful tools for generation, belief maintenance and experimentation. Here we specialize the use of compatible executable repairs to a finite decision interface and evaluate its representation, diagnostic access and execution feedback together.

The evaluation question is how to use remaining executable repair alternatives when the observations have not identified one model. Our interface lets each compatible program predict guard acceptance, keeps a weight on those predictions, and uses them to choose an execution route, buy a diagnostic observation, or take a backup. A point repair supplies one prediction; the retained belief also supplies disagreement that changes predicted route success and expected cost, the possible decision changes after a test, and candidates that ordinary execution can rule out.

\textbf{Choosing the design.} Table \ref{tab:methods} contrasts the closest approaches with the interface studied here by model or repair representation, source and use of uncertainty, and evaluated objective.

Each design choice below is paired with the comparison that tests it. Retention is tested by comparing the AI point repair, a single repair generated by a language model, with the AI-weighted mixture without tests (Table \ref{tab:data1}). The source of weight is tested with uniform and minimum-description-length (MDL) beliefs (Tables \ref{tab:data1} and \ref{tab:data2}). Observation reachability is tested by replacing flat test prices with reset and setup charges (Section 5.2). Whether diagnosis adds anything to what execution already teaches is tested by giving every repeated-work arm the same normal-execution updates, so that diagnosis competes with learning and not with a frozen error (Tables \ref{tab:data3} and \ref{tab:data5}). The selector itself is tested against random testing and information gain with the same four-test cap (Section 5.6). These comparisons identify the effects of specified interface choices.

Figure 1 shows the world and repeated-work decision loop; Figure 2 illustrates a recorded direct-access decision. The experiments yield three findings.

\begin{itemize}\setlength{\itemsep}{3pt}
\item \textbf{Mixing a generated repair with its compatible alternatives helps before any test is bought.} A fixed mixture of half AI repair and half uniform weight has lower aggregate independent-task cost than either its point-repair or its uniform endpoint. The paired device interval excludes zero for the contrast to point repair, but includes zero for the contrast to uniform retention, so the latter ordering is descriptive.
\item \textbf{Value-directed probes pay when information is inexpensive.} The benefit shrinks when reaching the test condition must itself be paid for, and it can disappear on isolated tasks.
\item \textbf{Repeated execution is itself a source of information.} A matched-budget comparison of workload-valued, random, and information-gain (EIG) test selection shows that workload-valued selection has lower aggregate cost than random selection under the AI-weighted belief. Its margin over EIG rests on two devices. Under uniform beliefs the order reverses and information gain is cheaper, and uniform normal learning without any tests already has a lower aggregate cost than AI-weighted workload-valued selection. A diagnostic method must therefore be evaluated together with repair weighting and against normal learning and other selectors, not only against a frozen erroneous model.
\end{itemize}
\suppressfloats[t]
\begin{table}[htbp]\centering\small\caption{Methodological comparison drawn from the cited sources; the last column gives the objective each study evaluated.}\label{tab:methods}

\begin{tabularx}{\linewidth}{XXXX}\toprule
Approach & Model or repair representation & Source and use of uncertainty & Evaluated objective and validation \\\midrule
Domain repair and action-model learning: FAMA \citep{ref11}, plan-based repair \citep{ref12}, counter-example repair \citep{ref13} & STRIPS preconditions and effects of a planning domain; grounded or lifted repairs & Consistency with partially observed traces or with positive and negative plans; repair sets are diagnoses, and a modeler may judge alternative repairs iteratively \citep{ref12} & Recovered model or cardinality-minimal or optimal repair; model-quality measures, plan validity, and repair benchmarks. Execution-time costs are not the evaluated quantity \\
Code world models: WorldCoder \citep{ref4}, GIF-MCTS \citep{ref5} & Python transition and reward programs & Replay data and an optimistic goal-reachability constraint \citep{ref4}; program-search values reflect predictive accuracy on stored transitions \citep{ref5}; acting supplies further data & Program that fits data and supports planning; GIF-MCTS compares with an earlier WorldCoder version \citep{ref5} \\
Hypothesis revision with experiments: LLM-SMC-S \citep{ref10} & Weighted natural-language rule particles revised by a language model & Bayesian posterior over particles; experiments chosen by information gain & Posterior prediction and informative experiments over language hypotheses; not costed route execution \\
This study & Finite executable guard repairs, as truth-table masks or syntax-counted disjunctive normal form (DNF) programs & Compatible repairs weighted by uniform, MDL-weighted, or AI-weighted mass and conditioned on every observation; predictions enter route, test, and backup choices; normal execution also updates the belief & Total task cost with priced tests, reset and setup access, and backup, on a fixed synthetic cohort with matched-belief controls \\
\bottomrule\end{tabularx}\end{table}

Our contribution is a reproducible decision interface for finite executable repairs, together with an analysis of three design choices: retaining compatible hypotheses, charging diagnostic access, and learning during ordinary work. It offers a concrete way to use generated repairs before they are uniquely identified. The experiments keep candidate generation apart from candidate use, and show that aggregate savings are concentrated on a subset of devices. The interface integrates Bayesian conditioning, expected-utility experimental design, of which information gain is one common instance \citep{ref14}, belief-state planning that joins acting and information gathering \citep{ref15}, and diagnosis from conflicting observations \citep{ref16} at the repair-to-execution boundary for finite executable guards, and measures when this integration pays.

\FloatBarrier
\section{The two-stage guard world}

Figure 1(a) illustrates the evaluated finite world. Model inputs are the structured guard observations, and all reported costs are computed from analytical execution records.

Each synthetic device consists of two sequential stages. A stage accepts or rejects a context, represented as a bit vector. A hidden reference guard determines acceptance, and only the evaluator reads it. The controller sees outcomes of actions it takes, never the guard itself. A task supplies a menu of context pairs \((q_A,q_B)\) to try. Stage B is executed only after stage A succeeds. An available backup completes the task at a known cost, so the controller always has a way to finish.

A repair program defines a candidate guard. Each model to be repaired starts from the same erroneous guard x0, which accepts exactly when Boolean input 0 is true; this input guard is supplied to AI generation. It is fixed independently of the hidden reference guard, whose construction defines the simulated world. The small grammar permits AND, OR, literal negation, and Boolean constants with at most four literal occurrences. Distinct candidate behaviors are represented as truth-table masks; multiple expressions with the same truth table share one candidate. Every observation removes candidates that disagree with it, so the set of compatible candidates shrinks as the controller works.

The finite reference guards act as hidden, static model parameters; the current menu and execution stage are observed state. A diagnostic action returns an acceptance or rejection, an execution action incurs route cost and supplies ordinary feedback, and the backup ends the task at a known cost. Augmenting the state with the model parameters gives a partially observable decision problem in the sense of belief-state planning, where actions both change the world and yield information \citep{ref15}. The controller approximates this problem with two assumptions: the two stage beliefs are maintained separately, and diagnostic tests are scored by one-step lookahead on expected task cost (Section 3.3). Ordinary execution supplies observations for the next belief update. Neither approximation carries an optimality guarantee for the full problem.

A concrete reading is a controller that must sequence two operations whose enabling conditions are incompletely known, as in a manufacturing interlock or a software workflow. The evaluated domain is the synthetic guard-controlled device. Context identifiers encode four Boolean settings without assigning them physical sensor semantics.

\FloatBarrier
\section{Method: repair beliefs and execution decisions}

\subsection{Hypothesis representation}

Let \(\mathcal H_j(D)\) denote compatible candidates for stage \(j\), and \(w_j(h)\) their normalized weights. The uniform variant assigns equal weight to each compatible behavior. A concentrated variant assigns half the probability mass uniformly and half to a selected compatible repair, obtained from either the AI (the AI-weighted belief) or a minimum-description-length (MDL) rule. The small-grammar MDL rule first minimizes literal count, then behavior changes from the original guard, then the numeric mask. The half-mass mixture is a fixed weighting choice, not a calibrated population estimate. Keeping every behavior that is consistent with the evidence follows the diagnosis tradition \citep{ref16}, and the repair literature selects cardinality-minimal or otherwise optimal repair sets among such alternatives \citep{ref12,ref13}; the MDL rule plays the role of a minimality preference among the retained behaviors, while the point repair of Table \ref{tab:data1} plays the role of a single recovered model as produced by the approaches in the first row of Table \ref{tab:methods}.

The probability of stage acceptance at context \(q\) is

\[
p_j(q)=\sum_{h\in\mathcal H_j(D)} w_j(h)h(q).
\]

The half-mass allocation is made once at initialization. A new observation conditions the weights by eliminating incompatible candidates and renormalizing; the mixture is not reset to half mass after each observation. The larger-grammar study uses a different prior over program syntax, counting multiple syntactic programs even when they represent the same behavior; its results are reported separately.

\subsection{Choosing an execution route}

Action cost in the small-grammar tasks is \(c(q)=1+\operatorname{popcount}(q)/4\), where \(\operatorname{popcount}(q)\) is the number of true settings in context \(q\), and the backup costs \(b=8\). Under the stage-factorized belief, the expected cost of trying a route and using the backup after failure is

\[
L_w(q_A,q_B)=c(q_A)+p_A(q_A)c(q_B)
       +b[1-p_A(q_A)p_B(q_B)].
\]

The controller minimizes this quantity over the menu and the immediate-backup option. Uncertainty therefore affects execution even when no diagnostic test is purchased. Because the backup is guaranteed in these synthetic tasks, final completion does not discriminate between policies. We instead report first-route success, which counts completion of both stages before fallback and excludes immediate backup. A successful route can cost more than a cheaper route on another menu, so success count alone does not rank cost performance.

\subsection{Purchasing a diagnostic observation}

Let \(V(w,T)\) be the minimum expected execution cost for task \(T\). For a candidate test \(a\), the myopic value of information (VOI), net of test cost, is

\[
G(a)=V(w,T)-\mathbb E_{y\sim w}[V(w\mid a,y,T)]-c_{\rm test}(a).
\]

This is a posterior-dependent expected-utility criterion of the kind defined in Bayesian experimental design, where information gain is one common utility and not the only one \citep{ref14}; here the utility is the negative task cost. Outcomes are deterministic given the hidden guard, but the guard is uncertain, so the predicted outcome before a test is still a distribution. This instantiation gives diagnostic outcomes the same task-cost objective as route selection.

The independent-probe implementation selects the test with largest positive value, conditions the belief on the observed result, and repeats up to two tests. It stops when no positive-value test remains. The access-aware variant also charges resetting and preparing the device to reach a test. In particular, testing stage B requires an available successful stage-A setup. The reset charge is 0.25. A stage-A test costs \(0.25+c(q)\); a stage-B test costs \(0.25+c(q_{\rm setup})+c(q)\). The setup is the cheapest context accepted by every surviving stage-A candidate, with context index breaking ties. If none exists, stage-B tests are unavailable. Because every surviving candidate accepts the setup context, setup observations are certain under the belief. Complete support here means retaining every grammar-consistent candidate. Setup observations therefore add no information; only the queried guard changes. These costs replace the flat per-query charges. 

The isolated access-aware comparison also includes entropy-per-cost selection. It ranks each feasible nondeterministic test by binary entropy of its predicted outcome divided by the full reset, setup and query cost above. It takes at most two tests per task, stopping when no feasible nondeterministic test remains; it does not require positive net task value. Ties use stage then context index, in increasing order.

For repeated workloads, the full remaining menu sequence is given to every method. Workload-valued selection sums the expected reduction over remaining task menus, uses a shared diagnostic budget of four tests per device, and conditions the belief on normal execution outcomes for every comparison method.

\begin{minipage}{\linewidth}
\begin{quote}\small\ttfamily\raggedright
\hspace*{0.0em}retain repairs compatible with the current observation history\par
\hspace*{0.0em}while a diagnostic budget remains:\par
\hspace*{0.5em}evaluate feasible tests using expected task-cost reduction minus test cost\par
\hspace*{0.5em}stop when the best net value is nonpositive\par
\hspace*{0.5em}execute that test and condition the repair belief\par
\hspace*{0.0em}choose the least expected-cost route or immediate backup\par
\hspace*{0.0em}execute; in the repeated-workload variant, condition on execution observations\par
\end{quote}
\end{minipage}

\subsection{Decision relevance and stopping}

The stopping criterion concerns a decision, not full model identification. For a fixed feasible route set and a test that only changes information, suppose one route minimizes expected cost under every possible test outcome. The expected cost of that route before the test equals its probability-weighted cost after the test, by conditioning. The test then has zero gross decision value and negative net value whenever its cost is positive. This elementary consequence of the decision rule explains why identifying every remaining guard detail is unnecessary, and it is the reason the interface can act before a unique repair is available. Conversely, different branch-optimal routes create an opportunity for a valuable test, but its expected benefit still has to exceed its access cost.

The criterion is myopic. Several jointly useful tests can each have insufficient one-step value, and workload-valued selection does not fully account for information that later normal execution will acquire. 

\subsection{A recorded decision}

A saved task from the small-grammar cohort (Section 4) shows the complete decision. For device 5 of Figure 4, menu 7, committing to the AI repair chooses immediate backup at cost 8. The AI-weighted belief also chooses backup without a test. At unit test cost, the controller estimates a net value of 0.276 for testing stage B in context 11. The independent direct test of context 11 is accepted, which changes the preferred route to contexts \((5,11)\). The route succeeds at execution cost 3.25, for a total of 4.25 including the probe. For comparison, the AI point-repair policy pays 8 for the backup, and the uniform policy without tests pays 3.25. The case comes from the independent-probe condition and was selected to expose the input, observation, and route change. The initial support contains 20 stage-A and nine stage-B repairs. Their weighted prediction assigns probability 0.278 to acceptance of B at context 11. Acceptance retains five B repairs and changes the best expected route cost from backup 8 to 3.41; the rejected-observation branch would leave best expected cost 8. Expected cost under remaining uncertainty and the realised successful execution cost 3.25 are different quantities. It uses direct access to the stage-B query; the access-aware condition instead charges reaching stage B (Section 3.3), so the 4.25 total is not comparable to the access-aware costs in Section 5.6.

\begin{figure}[htbp]\centering\includegraphics[width=\linewidth]{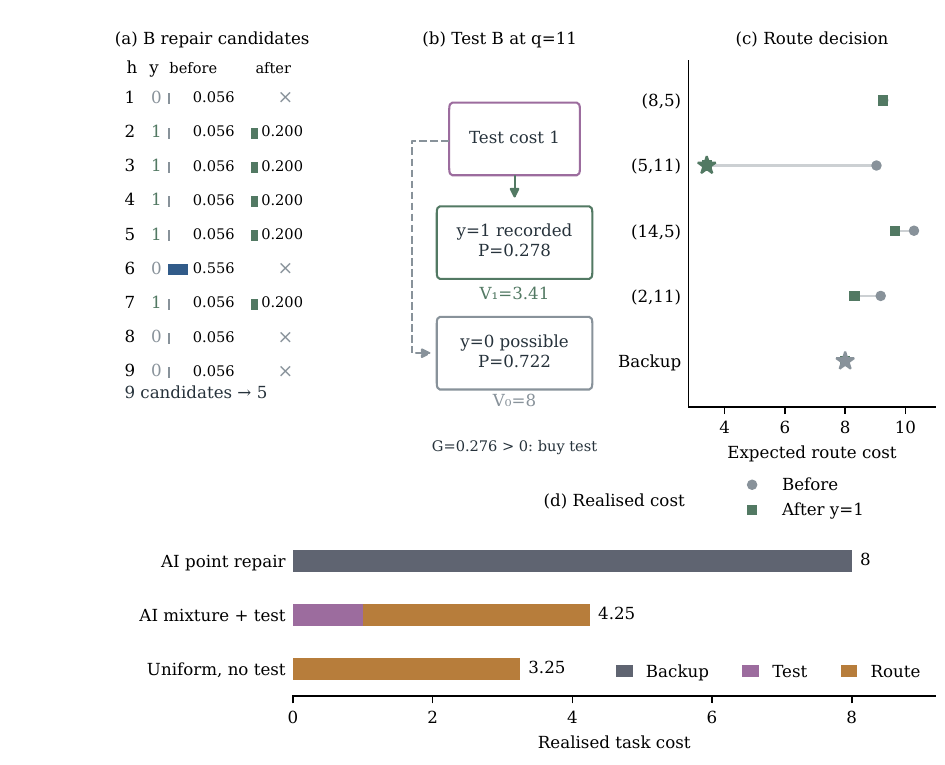}\caption{Repair-to-execution mechanism for one recorded task (device 5 of Figure 4, menu 7). Candidate h predicts query outcome y; bars show weights before and after acceptance (navy marks the AI repair). Observed acceptance conditions the belief and changes expected route costs; stars mark the chosen alternatives before (grey, Backup) and after the observation (green, (5,11)); × marks a candidate eliminated by the accepted observation. Panel (d) gives realised abstract task cost (backup cost 8), including the cheaper uniform no-test alternative. The query uses direct, unit-cost stage-B access. Routes are context pairs \((q_A,q_B)\), indexed 0–15. Stage A retains 20 candidates throughout this local stage-B test.}\end{figure}

\section{Experimental setting and accounting}

\textbf{Small-grammar cohort.} Each guard has four Boolean input variables. The cohort contains 48 guards arranged into 24 two-stage devices, with 64 task menus per device and four possible routes per menu. Guards have minimum literal counts of two, three, or four, with 16 cases per stratum. Six common observations precede decisions: contexts 0 and 15 and four tests maximizing binary outcome entropy under uniform weight over compatible small-grammar behaviors, chosen separately per guard (ties use the smaller context index). AI generation uses these observations to produce compatible repairs. The archived GPT-6-Astra high-reasoning outputs were generated in three batches of 16 cases, and shared generation batches are a source of dependence. Route menus use contexts held out from the initial observation history. The 1,536 task decisions are repeated measurements on 24 devices.

\textbf{Comparisons.} The controls follow the design choices listed in Section 1 and the rows of Table \ref{tab:methods}. The point repair and the AI-weighted mixture share the same generated repair and observations, testing replacement of a point prediction by the specified weighted candidate belief. The AI-weighted and MDL-weighted variants put half the initial mass on their selected repair and half uniformly on compatible behaviors; the uniform variant spreads all mass equally. The access-aware condition changes test feasibility and charges reset and setup as well as the query.

\textbf{Repeated work.} The repeated-workload study runs the same devices and menus in sequence. Every arm conditions its belief on the outcomes of ordinary execution; diagnostic arms add tests on top of these same updates. Section 5.6 compares complete selectors with a common four-test cap, separating their diagnostic choices from recovery of a frozen error.

\textbf{Larger grammar.} This group contains twelve eight-variable guards expressed in DNF with at most three clauses and three literals per clause, and six paired devices, yielding 384 menu decisions. It compares single repairs with full syntax-counted candidate predictions under saved observation histories. There are six initial observations and four additional tests per guard, with 64 common held-out contexts per world. Candidate counting includes the constant-true clause and accounts for syntactic multiplicity. This group isolates the prediction and route-selection component. The AI-selected history comprises the same six initial observations followed by four test contexts chosen by GPT-6-Astra with high reasoning effort from the allowed, non-holdout contexts, using its current observations and compatible DNF proposals. The exact syntactic-EIG history instead chooses four contexts by the acceptance/rejection split of the complete DNF syntax-counted candidate space, computed by inclusion–exclusion without enumerating all programs. Both histories use the same initial observations and test cap. The larger-grammar action cost is \(c(q)=1+\operatorname{popcount}(q)/8\), with backup cost 8; its task analysis adds no online diagnostic actions. The larger-grammar MDL point control selects the shortest consistent DNF found by bounded branch-and-bound and repeated greedy-cover search, retaining the first best expression on ties; implementation settings and source are supplied in the companion material.

\textbf{Cost accounting.} Reported cost is an abstract task objective. Tables \ref{tab:data1}, \ref{tab:data3}, and \ref{tab:data5} include charged diagnostic actions and execution; Table \ref{tab:data2} reports matched-belief differences under those same charges. Common initial data acquisition and model-generation latency (model computation time) are outside those totals, and the larger-grammar table reports execution cost only. These task costs measure action expenditure rather than end-to-end wall-clock time.

\textbf{Units and dependence.} The study is exploratory, with summaries and resampling intervals conditional on the saved cohort rather than population confidence intervals. Event summaries compare guard refutations and execution outcomes without estimating causal mediation. Leave-one-device-out ranges describe sensitivity to device composition. Concentration controls and access and workload variants reuse the small-grammar cohort, and the larger-grammar execution analysis reuses its prediction histories. The device pair is the aggregation unit for heterogeneous effects, and task menus are correlated within devices. Prior initial observations are common within matched comparisons. Resampling intervals use 20,000 paired device resamples; random test selection is repeated across 32 seeds. Seeds and implementation settings are given in the companion README.

\FloatBarrier
\section{Results}

\subsection{Repair uncertainty and test cost}

Mixing a generated repair with its compatible alternatives is useful before the model is recovered. AI generation fully restores 16 of the 48 guards, yet the AI-weighted belief without further tests costs less than committing to one AI repair, and inexpensive value-directed tests reduce cost further.

\begin{table}[htbp]\centering\small\caption{Small-grammar results over 1,536 menu decisions.}\label{tab:data1}

\begin{tabularx}{\linewidth}{Xrrr}\toprule
Decision rule & Cost & First-route success & Tests \\\midrule
AI point repair & 9227.25 & 681 & 0 \\
AI-weighted, no test & 9029.75 & 718 & 0 \\
AI-weighted, test cost 0.25 & 8322.5 & 899 & 1815 \\
AI-weighted, test cost 1 & 9018.75 & 745 & 139 \\
Uniform, no test & 9263.75 & 737 & 0 \\
Uniform, test cost 0.25 & 8377.75 & 909 & 2308 \\
Uniform, test cost 1 & 9131.5 & 823 & 591 \\
\bottomrule\end{tabularx}\end{table}

At test cost 0.25, the AI-weighted policy reduces its own no-test total from 9,029.75 to 8,322.5 (7.83 percent). Device-level totals improve for 18 of 24 devices and worsen for 6. The uniform policy reduces its own total from 9,263.75 to 8,377.75. At unit test cost these savings shrink markedly, and at cost 4 both policies buy no tests.

The AI-weighted versus MDL-weighted cost difference is concentrated in three devices. The MDL-weighted belief without tests costs 9,297.25, compared with 9,029.75 for the AI-weighted mixture. The no-test AI-versus-MDL reduction of 267.5 comes from two improving devices, one worsening device, and 21 ties. At unit test cost, the MDL total is 9,133.75 versus 9,018.75 for the AI-weighted mixture. At test cost 0.25, the MDL-weighted arm costs 8,346.75 with 1,823 tests and 896 first-route successes, versus 8,322.5 with 1,815 tests and 899 successes for the AI-weighted arm. The AI-minus-MDL total of -24.25 comes from three improving devices and 21 ties. The paired-device intervals in Table \ref{tab:data2} include or touch zero for every comparison.

\begin{table}[htbp]\centering\small\caption{AI-mixture cost minus the indicated belief, aggregated over the same 24 devices. Negative totals favor AI. Intervals summarize the mean device difference using 20,000 paired resamples; see Section 4 for their interpretation. }\label{tab:data2}

\begin{tabularx}{\linewidth}{Xrrr}\toprule
Diagnostic condition & Comparator & Total difference & Mean difference, 95\% interval \\\midrule
No test & Uniform & -234.00 & -9.75 [-26.43, 4.54] \\
No test & MDL-weighted belief & -267.50 & -11.15 [-41.31, 7.88] \\
Test cost 0.25 & Uniform & -55.25 & -2.30 [-6.81, 3.51] \\
Test cost 0.25 & MDL-weighted belief & -24.25 & -1.01 [-2.84, 0.00] \\
Test cost 1 & Uniform & -112.75 & -4.70 [-14.78, 5.44] \\
Test cost 1 & MDL-weighted belief & -115.00 & -4.79 [-16.20, 2.40] \\
\bottomrule\end{tabularx}\end{table}

\begin{figure}[htbp]\centering\includegraphics[width=\linewidth]{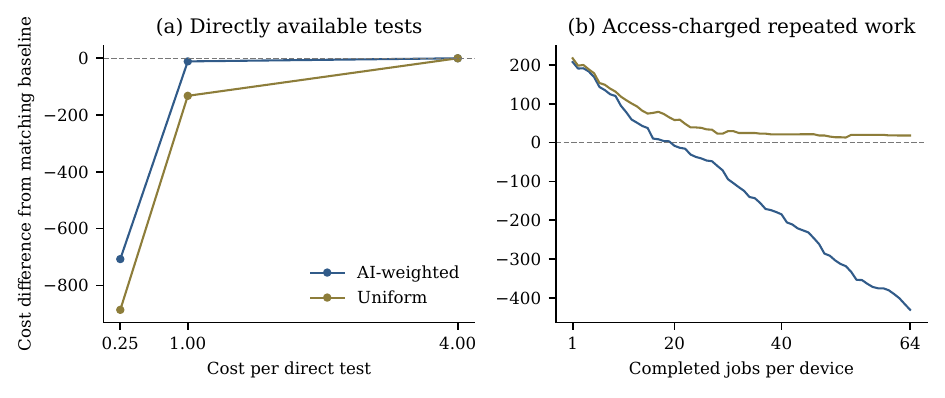}\caption{Diagnostic cost differences from the matching no-test policy. Panel (a) aggregates 1,536 separately initialized menu decisions on 24 devices; panel (b) shows cumulative differences over 64 sequential jobs, with normal-execution updates and at most four access-charged tests per device (reset and setup included). Panel (a) lines join three tested cost levels. Negative values indicate lower total cost. Curves are observed cohort aggregates, without uncertainty intervals. Colors in (b) follow (a); the initial positive difference includes diagnostic spending before the first job.}\end{figure}

\subsection{Access-charged diagnosis in repeated workloads}

With reset and setup charges, value-directed testing no longer lowers isolated-task cost. On isolated tasks, the access-aware uniform VOI policy costs 9,278 versus 9,263.75 without diagnosis, and the AI-weighted policy purchases no tests. Entropy-per-cost testing costs 13,672.5 with the AI-weighted belief and 13,836.5 with the uniform belief. Information that is useful when directly queryable can be uneconomic once reset and preparation are included.

A repeated workload can amortize diagnostics. Table \ref{tab:data3} totals all 64 jobs on each of the same 24 devices, and every arm learns from ordinary execution.

\begin{table}[htbp]\centering\small\caption{Repeated work, 24 devices, 64 jobs each.}\label{tab:data3}

\begin{tabularx}{\linewidth}{Xrrr}\toprule
Decision rule & Cost & First-route success & Tests \\\midrule
AI-weighted, normal learning & 8500.25 & 762 & 0 \\
AI-weighted, workload-valued selection & 8069.25 & 885 & 96 \\
MDL-weighted belief, normal learning & 8502.5 & 761 & 0 \\
MDL-weighted belief, workload-valued selection & 8070.5 & 885 & 96 \\
Uniform, normal learning & 7959.5 & 886 & 0 \\
Uniform, workload-valued selection & 7977.75 & 913 & 96 \\
\bottomrule\end{tabularx}\end{table}

Workload-valued selection lowers the AI-weighted total by 431 (5.07 percent), from 8,500.25 to 8,069.25, while first-route successes increase from 762 to 885. The aggregate difference first becomes nonpositive after 20 jobs per device and remains nonpositive through job 64 (Figure 3b). Savings occur on 11 devices and extra cost on 13, so the aggregate gain comes from larger reductions on a subset rather than uniform small improvements. The MDL-weighted counterpart reaches almost the same cost, reducing its normal-learning baseline of 8,502.5 to 8,070.5.

Uniform normal learning has the lowest total among the Table \ref{tab:data3} arms, 109.75 (1.36 percent) below AI-weighted workload-valued selection. Adding workload-valued selection to the uniform belief raises its cost to 7,977.75 even though first-route successes rise. The workload policy uses all four permitted probes on every device; Section 5.6 compares it with same-budget random selection.

\subsection{Device-level heterogeneity and execution events}

With paired device resampling, the 95 percent percentile intervals for mean cost differences per device are -17.58 to -0.77 for the AI-weighted mixture minus AI point repair; -44.28 to -14.94 for cheap tests minus the same mixture without tests; and -34.26 to -3.93 for workload-valued selection minus the same AI-weighted belief with normal learning. Omitting each batch in turn preserves the direction of all three aggregate differences, with workload savings ranging from 4.47 to 6.14 percent of the matching baseline. Comparisons with uniform normal learning separate two gaps. AI-weighted normal learning costs 540.75 more in total (mean +22.53 per device; resampling interval 9.80 to 36.49), and workload-valued selection recovers 431 of that difference. The remaining 109.75 (mean +4.57 per device) leaves uniform normal learning with the lower total on this cohort, while its resampling interval for the mean, -2.97 to 12.64, includes zero.

The paired distribution shows that aggregate savings do not describe a typical device. The no-test mixture beats the AI point repair on 8 devices, ties on 7, and loses on 9. Workload-valued selection improves 11 devices and worsens 13; its median cost difference is +4.375 even though the mean is negative. Figure 4 shows every device. Leaving out any one device preserves a negative mean cost difference: -9.00 to -5.38 for mixture versus point repair, and -19.25 to -13.77 for workload-valued selection versus AI-weighted normal learning.

\begin{figure}[htbp]\centering\includegraphics[width=\linewidth]{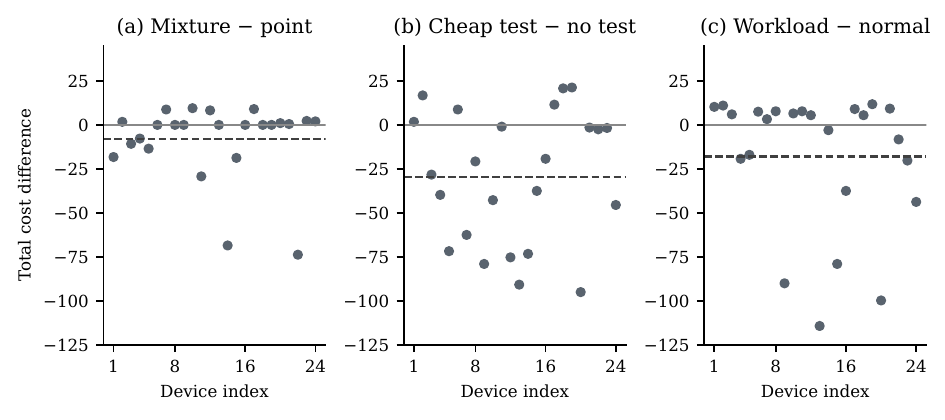}\caption{Device-paired total-cost differences for the fixed cohort. Each dot is one device’s total over 64 menus: separately initialized decisions in (a,b), direct tests at cost 0.25 in (b), and sequential work with access-charged tests in (c). Negative values favor the first policy. The dashed line is the sample mean. The three panels show mixture versus point repair, cheap tests versus the matching no-test mixture, and workload-valued selection versus the matching AI-weighted normal-learning baseline.  Each device index identifies the same device in all three panels.}\end{figure}

Saved execution events show what the diagnostics change. AI-weighted normal learning chooses immediate backup on 741 jobs and receives 1,567 normal-execution observations; adding workload-valued selection changes these totals to 627 and 1,798. Uniform normal learning chooses backup on 571 jobs and obtains 1,880 observations. Of 32 initially incorrect AI guards, observations refute 16 under AI-weighted normal learning, 25 with diagnostics, and 30 under uniform normal learning. The 1,798 observations exclude all 96 diagnostic probes and their setup and reset actions. The policies share 16 refuted guards; nine are refuted only with diagnostics, and none only with AI-weighted normal learning. All nine additional guards are first refuted by diagnostic probes. Of the 25 total first refutations in that policy, 20 occur in probes and five in normal execution. Eight of the nine additional guards belong to the cost-benefiting devices, versus one in the other devices.

Intervention thus changes which mistakes become observable. Every device receives normal observations. Mean baseline backup counts are 30.18 in the benefiting group and 31.46 in the other group; the groups differ in which guards the probes refute. 

\subsection{Candidate retention in a larger grammar}

In the larger grammar, the AI-selected observation history followed by a single AI repair predicts 568 of 768 held-out outcomes correctly. Retaining the full syntax-counted belief on that same history yields 599 correct predictions. This comparison separates the value of the observations from the decision to collapse them into one repair.

\begin{table}[htbp]\centering\small\caption{Larger grammar, 384 menu decisions, execution cost only.}\label{tab:data4}

\begin{tabularx}{\linewidth}{Xrr}\toprule
Decision rule & Cost & First-route success \\\midrule
AI-selected history, point repair & 3106.25 & 42 \\
AI-selected history, MDL repair & 3081.625 & 49 \\
AI-selected history, full syntax-counted belief & 3054.625 & 51 \\
Exact syntactic-EIG history, full syntax-counted belief & 3061.125 & 44 \\
Immediate backup & 3072 & 0 \\
\bottomrule\end{tabularx}\end{table}

On the same AI-selected history, posterior-based route selection reduces execution cost from 3,106.25 for the AI point repair to 3,054.625, and first-route successes rise from 42 to 51. Against stronger references, the full belief costs 17.375 less than always using the backup and 6.5 less than exact syntactic-EIG observations followed by posterior decisions. Across the six devices, the full syntax-counted belief lowers cost on three, ties on one, and raises it on two. The six-device paired-bootstrap 95 percent interval for mean cost change is -19.63 to +1.52, which includes zero. Prediction improvement combines candidate retention with the syntax-counted weighting scheme; these saved comparisons do not isolate weighting from retention.

\subsection{Gap to the full-information oracle}

The full-information oracle knows the reference guards without paying for information, so it is an evaluation bound rather than an available controller. It costs 7,542.75 across the 1,536 small-grammar menus; immediate backup costs 12,288. The cheap-test AI-weighted mixture at 8,322.5 closes 47.56 percent of its no-test baseline's 1,487-unit oracle gap. In repeated work, the same oracle total applies to the same menus: AI-weighted workload-valued selection closes 45.01 percent of its normal-learning baseline's 957.5-unit gap, while uniform normal learning leaves a smaller gap of 416.75.

For the larger grammar, oracle cost is 2,603.25. Immediate backup is 468.75 units above that oracle; the full syntax-counted belief closes 17.375 of this gap (3.71 percent), leaving 451.375. The observed reduction is 10.26 percent of the AI point repair's oracle gap. 

\subsection{Matched-budget diagnostic selection}

The selection experiment extends the same 24-device, 64-job cohort to compare four ways of choosing diagnostics: workload-valued selection, random testing, maximum expected information gain (EIG), and no-test normal learning. Both the AI-weighted and uniform beliefs use the same candidate support, initial history, menu sequence, route rule, observation updates, and access, reset, and backup accounting. The workload and EIG selectors can read the same future menus, but EIG scores the diagnostic observation alone rather than its task-cost effect. The two selectors therefore differ in both use of future menus and scoring objective, and the comparison evaluates complete selectors without separating these components. For noiseless binary outcomes, the expected entropy reduction that EIG maximizes is the binary entropy of predicted acceptance, taken without dividing by cost, which differs from the earlier entropy-per-cost arm. All testing policies have the same limit of four tests per device and exclude inaccessible or already determined queries. Deterministic ties use stage and context order. Random testing samples uniformly among the valid queries across 32 seeds, paired across beliefs within each device.

\begin{table}[htbp]\centering\small\caption{Matched-budget selection on the 24-device, 64-job cohort.}\label{tab:data5}

\begin{tabularx}{\linewidth}{Xrrr}\toprule
Belief and selector & Total cost & Diagnostic cost & First-route success \\\midrule
AI-weighted, normal learning & 8500.25 & 0 & 762 \\
AI-weighted, workload-valued selection & 8069.25 & 224.5 & 885 \\
AI-weighted, information gain & 8216.25 & 218 & 859 \\
AI-weighted, random (mean) & 8346.97 & 228.09 & 834.13 \\
Uniform, normal learning & 7959.5 & 0 & 886 \\
Uniform, workload-valued selection & 7977.75 & 236.75 & 913 \\
Uniform, information gain & 7920.5 & 222.25 & 922 \\
Uniform, random (mean) & 8060.05 & 228.09 & 902.41 \\
\bottomrule\end{tabularx}\end{table}

All testing policies used 96 tests across the cohort, so cost differences reflect which tests were chosen at a fixed test count. Every device exhausts the four-test cap, so the nonpositive-value stopping rule never binds in this cohort. The investment is made almost entirely before work starts: in the AI-weighted condition, 95 of the 96 workload-selected tests occur before the first job and one before the second; EIG and random testing use all 96 before the first job; and all uniform-belief tests precede the first job. The comparison therefore concerns an initial diagnostic investment followed by normal learning, not distributed diagnostic scheduling. Actual diagnostic expenses differ because the chosen contexts and setup actions differ. Random rows are means across 32 selection seeds on the fixed worlds; deterministic rows are single policy runs. 

Under the AI-weighted belief, workload-valued selection costs 8,069.25, compared with 8,346.97 for mean random selection and 8,216.25 for EIG. These are reductions of 3.33 and 1.79 percent. The first is evidence of a selection effect beyond a same-budget random intervention. Workload-valued selection improves on the mean random cost on 16 of 24 devices and on EIG on 15; eight and nine devices respectively become more costly. Its aggregate cost is below each of the 32 random-seed totals, which range from 8,141.75 to 8,532.75 (standard deviation 89.66).

The margin over EIG depends on two devices. Paired device resampling gives a 95 percent interval of [-21.68, -2.22] for mean workload-minus-random cost under the AI-weighted belief, but [-21.11, 6.20] for workload-minus-EIG. Omitting the two largest workload-benefiting devices changes the remaining total EIG difference from -147 to +61. 

\textbf{Selector ordering under uniform beliefs.} There, EIG costs 7,920.5, workload-valued selection 7,977.75, and mean random selection 8,060.05. Workload-valued selection costs 57.25 more than EIG, with eight devices improving, two tying, and fourteen worsening; the paired interval for uniform workload-minus-EIG is [0.31, 4.52]. Uniform EIG costs 39 units (0.49 percent) less than uniform normal learning, a small ordering whose interval, [-7.72, 3.14], includes zero. Uniform random totals range from 7,957 to 8,254.75 (standard deviation 65.78), and workload-valued selection is below 29 of the 32 random totals. The 32-seed whiskers in Figure 5 describe random selection variability on the same worlds; the common absolute cost axis compares both beliefs directly.

\begin{figure}[htbp]\centering\includegraphics[width=\linewidth]{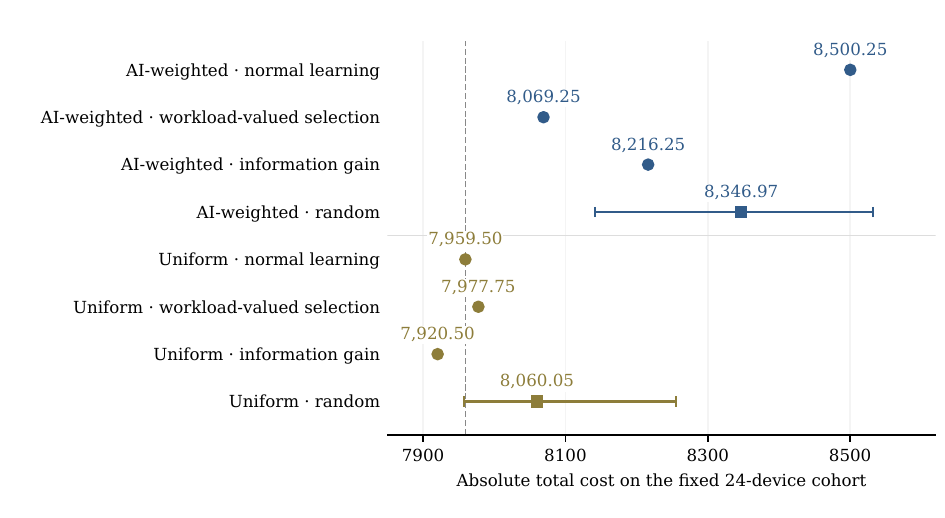}\caption{Absolute total cost of all eight belief–selector combinations on the fixed 24-device cohort. Circles show deterministic policies; squares and whiskers show random-selector means and ranges across 32 seeds. The dashed reference is uniform normal learning. All marks include test, setup, reset, execution, and fallback costs.}\end{figure}

The saved execution events show different backup and observation counts. Under the AI-weighted belief, workload-valued selection has fewer immediate backups and more normal-execution observations than the alternative selectors: 627 immediate backups and 1,798 normal observations, compared with EIG's 651 and 1,757 and the random means of 674.25 and 1,705.63. EIG and workload-valued selection nevertheless both refute 25 initially incorrect guards, so refutation count alone does not explain their cost ordering. Under uniform beliefs, EIG yields 1,904 normal observations and 30 refuted guards, versus workload-valued selection's 1,880 and 29. These event counts suggest that belief concentration, selected test contexts, and subsequent execution feedback interact. 

\FloatBarrier
\section{Related work}

\textbf{Model repair and acquisition.} The survey of Bercher, Sreedharan, and Vallati \citep{ref17} distinguishes learning a model from scratch from repairing a flawed one, organizes repair approaches by the constraint the repaired model must satisfy, such as making problems solvable or including or excluding given plans, and discusses which repaired model to aim for. We use that organization to place the present work. FAMA \citep{ref11} learns STRIPS action models by compiling partially observed plan traces, and optionally a partial action model, into classical planning, and it also supplies syntactic and semantic model-quality evaluation. Lin, Grastien, and Bercher \citep{ref12} repair a flawed domain so that a known-valid plan becomes executable, returning a cardinality-minimal repair, and the authors describe a modeler judging alternative repairs iteratively; repair alternatives are therefore part of that setting, and what is evaluated is the repair optimization. Lin et al. \citep{ref13} require both positive plans to be executable and counter-example plans to be invalid. Our accepted and rejected guard outcomes are simpler structured evidence of the same two kinds. The recent preprint of Karimi Bavandpour and Bercher \citep{ref18} evaluates open-weight language models on repair against such constraints and reports model-recovery F1 and test-pass validity. These metrics score a single generated model, whereas our task costs score decisions taken while several repairs remain compatible. Gragera et al. \citep[pp.344–346]{ref1} extend an existing self-repair compilation to use partially observed actions. Their experiments compare repaired Planning Domain Definition Language (PDDL) domains on goal solvability, observation compliance, and recovery of missing effects, against the FAMA \citep{ref11} action-model learning baseline and HIT, the hitting-set model repair of \citep{ref12}, as reported in that study. Our setting instead holds the compatible guard repairs as an execution-time belief and asks whether another observation changes route, diagnostic, or backup choice.

\textbf{Bayesian exploration and decision-aware models.} Dearden, Friedman, and Andre \citep[Sections 3–5]{ref2} maintain uncertainty over MDP transition and reward parameters, propagate it to Q-values, and combine expected value with a myopic perfect-information bonus. We instead condition a finite set of executable guard repairs and evaluate the actual binary outcomes of available diagnostic tests rather than idealized perfect information. Farahmand, Barreto, and Nikovski \citep[Section 2]{ref3} introduce a value-aware transition-model loss. Our controller leaves generated repairs intact and changes how their predictions enter downstream decisions. Modern Bayesian experimental design \citep{ref14} treats expected information gain as one common utility among general posterior-dependent expected utilities and covers adaptive designs; our workload-valued score uses expected task cost as that utility, and the EIG selector in Section 5.6 uses information gain.

\textbf{Executable world models.} WorldCoder \citep[Sections 2.2–2.5, Algorithm 1]{ref4} seeks an observation-consistent model satisfying an optimistic goal-reachability constraint, then plans with the current transition and reward program; acting supplies further data, so learning and acting already interact there. Its population of past programs supports refinement search. Our interface uses weighted predictions across compatible repairs in route and diagnostic decisions. GIF-MCTS \citep[Sections 3–4]{ref5} searches a tree of programs using generation, improvement, and bug-fixing actions scored against recorded transitions, and its planning evaluation loads a selected generated program. Those search populations serve a different role from an execution-time repair belief. The methods can be complementary, since a code generator could supply proposals to a belief-based controller. Our bounded guard grammar permits complete candidate conditioning and controlled comparisons between point-repair and belief-based execution policies.

\textbf{Diagnosis, interrogation, and decision-directed testing.} Reiter's theory of diagnosis \citep{ref16} characterizes minimal explanations of conflicting observations, and the hitting-set repair of \citep{ref12} is connected to that lineage. Heckerman, Breese, and Rommelse \citep[Sections 1–3]{ref6} minimize expected troubleshooting cost using observations, repairs, and a service fallback; our route, test, and backup choices apply the same expected-cost reasoning to retained executable repairs. Golovin, Krause, and Ray \citep[Section 3]{ref7} study identifying a hypothesis equivalence class and introduce an edge-cutting objective with approximation guarantees; the edge-cutting objective has a different structure from the myopic route-cost score used here. Verma, Marpally, and Srivastava \citep[Algorithms 1–2]{ref8} retain possible relational action models and generate queries that distinguish them. Their model-identification objective differs from choosing a useful route while ambiguity remains. These identification and troubleshooting objectives locate the repair-to-execution interface: remaining candidate uncertainty enters a useful action decision before full identification.

\textbf{Belief updating during action.} Belief-state planning in partially observable domains \citep{ref15} already joins reward and state-changing effects with information acquired through action, and BAMCP \citep[Algorithm 1]{ref9} samples models from a belief and searches over histories, accounting for information gained through action. Our workload heuristic instead scores diagnostics against remaining menus without forecasting subsequent passive learning. Piriyakulkij et al. \citep[Sections 2.1–2.2]{ref10} maintain weighted natural-language hypotheses with LLM-guided sequential Monte Carlo and select experiments by information gain, combining LLM hypotheses, posterior prediction, and experimentation. Our implementation enumerates finite executable guards, then charges route, fallback, and state-dependent diagnostic access in a task-cost objective. Section 5 evaluates this integration through controlled point-repair, belief, and diagnostic-policy comparisons.

\FloatBarrier
\section{Discussion}

\textbf{Information sources.} Diagnostic tests can change a route decision, but their value shrinks when reset and setup must be purchased (Sections 5.1--5.2). Ordinary execution supplies another source of evidence. In the repeated-work comparison, the broader uniform belief falls back less often, observes more outcomes and refutes more initially incorrect AI guards (Section 5.3). These event summaries locate differences between the policies; they do not identify the causal contribution of each event type.

\textbf{Design implications.} The matched comparisons locate benefits at the repair-to-execution boundary rather than in prediction accuracy or final completion alone. The larger-grammar study tests this distinction under identical histories (Section 5.4). The cheaper diagnostic selector changes with the belief weighting (Section 5.6), making belief concentration a candidate observable for choosing between task-cost and information-gain scoring, not a validated switching rule. Selection and weighting should therefore be evaluated together, against a controller that is already learning. In terms of Table \ref{tab:methods}, retained repairs support execution before unique model identification, while useful diagnostics depend on reachable observations and on how the repairs are weighted.

\textbf{Conditions of the evidence.} The quantitative findings concern deterministic finite guards, known task menus and costs, and complete candidate support within the chosen grammar. On 48 small-grammar guards from three generation batches, the AI-weighted and MDL-weighted belief variants use a half-mass mixture fixed at initialization, with stage-factorized beliefs and one-step test values. Uniform beliefs and point repairs supply the comparison endpoints. The larger-grammar full-belief variants use the syntax-counted prior on twelve guards. Costs follow the accounting in Section 4 and exclude initial acquisition and model inference. Applying the interface requires an executable candidate representation, observation likelihoods and known action costs.

\FloatBarrier
\section{Reproducibility}

\textbf{Generation provenance.} The three archived generation batches report the provider model identifier \texttt{gpt-6-astra} with high reasoning effort. Complete prompts, observation packets, output schemas, hashes and metadata accompany the fixed generated outputs in the arXiv ancillary files. The model snapshot and decoding settings are unavailable; the supplied generated outputs are fixed inputs to the reported comparisons.

\textbf{Companion material.} The ancillary archive contains the guard parser, planner, belief update and test-selection functions, together with saved tasks, initial observations, generated repairs and evaluator guards. It supplies replay scripts for the principal small-grammar comparisons and the MDL cheap-test arm, plus event-based checks for the selection comparison. Other small-grammar arms and larger-grammar analyses are supplied as saved decisions, predictions, histories and analysis scripts. These checks verify costs, belief support and event transitions on the supplied worlds.

\FloatBarrier
\section{Conclusion}

Retaining compatible repairs makes uncertainty available to execution decisions before a model is uniquely identified. We implemented this interface with expected-cost route selection, optional diagnostic tests, access charges, and updates from normal execution. Inexpensive tests reduce the AI-weighted policy's cost by 7.83 percent on the 24-device cohort. Access-charged workload-valued selection reduces its own baseline by 5.07 percent and has lower fixed-cohort cost than matched-budget random testing under the same AI-weighted belief (3.33 percent). Its 1.79 percent advantage over EIG is sensitive to two devices. With uniform beliefs the ordering reverses: EIG is cheaper than workload-valued selection and has the lowest total cost of the eight belief–selector combinations in Table \ref{tab:data5}, and uniform normal learning without tests has a lower total than every AI-weighted arm. In the larger grammar, the full syntax-counted belief predicts more held-out outcomes correctly (599 versus 568 of 768) than a single repair under the same history; its lower aggregate execution cost has an interval that includes zero. These results provide cohort evidence for mixing a generated repair with compatible alternatives rather than committing to it, pricing diagnostic access, and retaining information from ordinary execution.

\paragraph*{Use of AI assistants.} Under the authors' direction, AI assistants (OpenAI Codex and Anthropic Claude) assisted in implementing analysis code and in drafting and editing the manuscript text. The authors defined the research question, design, analysis and conclusions, verified all reported results, and take full responsibility for the content.

\bibliographystyle{unsrtnat}
\bibliography{references}
\end{document}